%% file: Thumm_2022_Provably_Safe_Deep_Reinforcement_Learning_for_Robotic_Manipulation_in_Human_Environments.tex
\documentclass[letterpaper, 10 pt, conference]{ieeeconf}  

\IEEEoverridecommandlockouts                              % This command is only needed if 
\usepackage{graphicx} % for pdf, bitmapped graphics files
\usepackage{times} % assumes new font selection scheme installed
\usepackage{amsmath} % assumes amsmath package installed
\usepackage{amssymb}  % assumes amsmath package installed
\usepackage{siunitx} 
\usepackage{tikz}
\usepackage[ruled, vlined, noend, linesnumbered]{algorithm2e}
\usepackage{cite}
\usepackage{hyperref}
\usepackage[capitalise]{cleveref}
\usepackage{tikz}
\usepackage{pgfplots}
\pgfplotsset{compat=1.5.1,}
\usepackage{pgfplotstable}
\usepackage{standalone}
\usepgfplotslibrary{fillbetween}
\usepackage[caption=false]{subfig}
\usepackage{booktabs}

\makeatletter
\newcommand{\removelatexerror}{\let\@latex@error\@gobble}
\makeatother

\definecolor{TUMblue}{rgb}{0.00, 0.40, 0.74}
\definecolor{TUMdarkblue}{rgb}{0.00, 0.32, 0.576}
\definecolor{TUMlightblue}{rgb}{0.392, 0.627, 0.784}
\definecolor{TUMlighterblue}{rgb}{0.596, 0.776, 0.917}
\definecolor{TUMgray}{rgb}{0.6, 0.6, 0.6}
\definecolor{TUMlightgray}{rgb}{0.855, 0.843, 0.796}
\definecolor{TUMorange}{rgb}{0.89, 0.447, 0.133}
\definecolor{TUMgreen}{rgb}{0.635, 0.678, 0.00}

\definecolor{mygreen}{RGB}{0, 146, 0}
\definecolor{mygray}{rgb}{0.85, 0.85, 0.85}

\usetikzlibrary{arrows}
\usetikzlibrary{positioning}
\tikzstyle{gray_box}=[draw, fill=TUMgray, minimum size=2em]
\tikzstyle{blue_box}=[draw, fill=TUMblue, text=white, minimum size=2em]

\SetCommentSty{mycommfont}
\def\CC{{C\nolinebreak[4]\hspace{+.05em}\raisebox{.1ex}{\small\bf ++~}}}
\def\etal{~et~al.}

\title{\LARGE \bf
Provably Safe Deep Reinforcement Learning for Robotic Manipulation in Human Environments
}

\author{Jakob Thumm and Matthias Althoff% <-this % stops a space
\thanks{The authors are with the Department of Informatics, Technical University of Munich, 85748 Garching, Germany.
        {\tt\small jakob.thumm@tum.de}, {\tt\small althoff@tum.de}}%
}
\begin{document}

%%%%%% Abbreviations %%%%%%%%%%%
% RL = Reinforcement learning
% LTT = Long-term trajectory
% SAC = Soft-Actor Critic
% HER = Hindsight Experience Replay
% DoF = Degree of freedom

\maketitle
\thispagestyle{empty}
\pagestyle{empty}

%%%%%%%%%%%%%%%%%%%%%%%%%%%%%%%%%%%%%%%%%%%%%%%%%%%%%%%%%%%%%%%%%%%%%%%%%%%%%%%%
\input{abstract}

%%%%%%%%%%%%%%%%%%%%%%%%%%%%%%%%%%%%%%%%%%%%%%%%%%%%%%%%%%%%%%%%%%%%%%%%%%%%%%%%
\section{INTRODUCTION} \label{Sec_Intro}
    \input{introduction}

\section{PRELIMINARIES} \label{Sec_Preliminaries}
    \input{preliminaries}

\section{METHODOLOGY} \label{Sec_Methods}
    \input{methods}

\section{RESULTS} \label{Sec_Results}
    \input{results}

\section{CONCLUSIONS} \label{Sec_Conclusion}
    \input{conclusion}

\section*{ACKNOWLEDGMENT}
The authors gratefully acknowledge financial support by
the Central Innovation Programme of the German Federal Government under grant ZF4086004LP7 and the Horizon 2020 EU Framework Project CONCERT under grant 101016007.

\addtolength{\textheight}{-5.8cm}   % This command serves to balance the column lengths

%%%%%%%%%%%%%%%%%%%%%%%%%%%%%%%%%%%%%%%%%%%%%%%%%%%%%%%%%%%%%%%%%%%%%%%%%%%%%%%%

%%%%%%%%%%%%%%%%%%%%%%%%%%%%%%%%%%%%%%%%%%%%%%%%%%%%%%%%%%%%%%%%%%%%%%%%%%%%%%%%

%%%%%%%%%%%%%%%%%%%%%%%%%%%%%%%%%%%%%%%%%%%%%%%%%%%%%%%%%%%%%%%%%%%%%%%%%%%%%%%%
%\section*{APPENDIX}
%\renewcommand{\arraystretch}{1.5}

%%%%%%%%%%%%%%%%%%%%%%%%%%%%%%%%%%%%%%%%%%%%%%%%%%%%%%%%%%%%%%%%%%%%%%%%%%%%%%%%

\bibliographystyle{IEEEtran}
\bibliography{library}

\end{document}

%% file: abstract.tex
\begin{abstract}
Deep reinforcement learning (RL) has shown promising results in the motion planning of manipulators.
However, no method guarantees the safety of highly dynamic obstacles, such as humans, in RL-based manipulator control.
This lack of formal safety assurances prevents the application of RL for manipulators in real-world human environments.
Therefore, we propose a shielding mechanism that ensures ISO-verified human safety while training and deploying RL algorithms on manipulators.
We utilize a fast reachability analysis of humans and manipulators to guarantee that the manipulator comes to a complete stop before a human is within its range.
Our proposed method guarantees safety and significantly improves the RL performance by preventing episode-ending collisions.
We demonstrate the performance of our proposed method in simulation using human motion capture data.
\end{abstract}
% Word count: 126

%% file: introduction.tex
In recent years, researchers solved many complex manipulation tasks using deep reinforcement learning (RL), such as operating door handles~\cite{gu_2017_DeepReinforcement}, playing table tennis~\cite{gao_2020_RoboticTable}, stacking boxes~\cite{li_2020_PracticalMultiObject}, and controlling multiple robotic arms~\cite{prianto_2020_PathPlanning}.
Furthermore, recent work~\cite{el-shamouty_2020_SafeHumanRobot, pham_2018_OptLayerPractical, sangiovanni_2021_SelfConfiguringRobot} demonstrated that RL-controlled manipulators could successfully maneuver in environments with dynamic obstacles, dodging moving obstacles, and reaching the goal state consistently.
Despite these promising results, one open challenge in RL is to formally guarantee the safety of surrounding humans due to their unpredictable movement and many degrees of freedom (DoF).
%So far, no RL approach provides formal safety guarantees for robotic manipulators in dynamic environments without a piori known obstacle kinematics.
%Especially the unpredictable movement behavior of humans and their many degrees of freedom (DoF) prevents the application of common RL methods in human environments.
The lack of safety assurances makes applying any vanilla RL method in human working environments irresponsible.

\begin{figure}[t]
\centering
\vspace{0.17cm}
\includegraphics[width=\columnwidth]{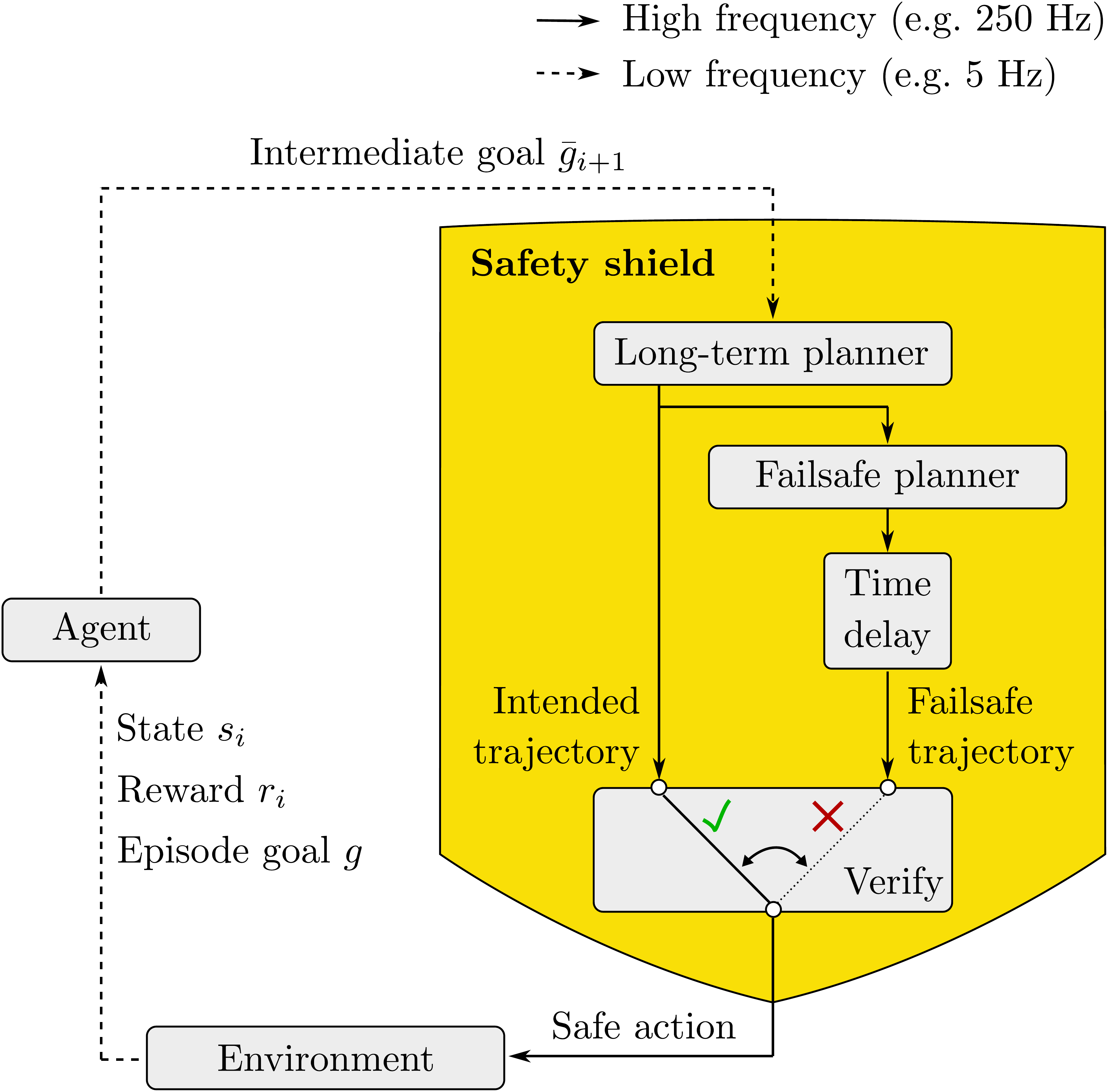}
\caption{Structure of the proposed RL agent with safety shield.}
\label{fig:concept}
\end{figure}

To overcome these restrictions, we propose a \textit{safety shield} concept for RL that guarantees human safety at all times, as shown in~\cref{fig:concept}.
In essence, our safety concept is \textit{speed and separation monitoring} (SSM) according to DIN~EN~ISO~10218--1~2021, 5.10.3~\cite{_2021_RoboticsSafety} warranting that the robot is in a completely stopped state before any collision with a human could occur.
We need to verify the robot trajectory at a high frequency to achieve this strong safety criterion while still being able to maneuver close to humans.
However, for the RL agent it suffices to output actions at a low frequency to make long-term decisions like moving around a human to reach its goal.
Our safety shield combines the low-frequency RL agent with our high-frequency formal verification. 
The RL agent used in this work is a state-of-the-art soft actor-critic (SAC)~\cite{haarnoja_2018_SoftActorCritic} with hindsight experience replay (HER)~\cite{andrychowicz_2017_HindsightExperience}.
However, our safety shield can be used with any online RL agent.
This work presents the first provably safe robot manipulator control based on deep RL in human environments.

\subsection{Related work}
Gu\etal~\cite{gu_2017_DeepReinforcement} were one of the first to show that various complex manipulation tasks can be learned using only off-policy deep RL methods. 
They presented a method to train multiple real robots in parallel to solve a challenging door-opening task.
Shortly later, \cite{andrychowicz_2017_HindsightExperience} proposed HER to achieve randomized manipulation goals.
The combination of HER with off-policy RL contributed to many remarkable results in pushing, pick and place, and throwing tasks as presented in~\cite{andrychowicz_2017_HindsightExperience, ren_2019_ExplorationHindsight, bing_2021_ComplexRobotic}.
Recently, HER was combined with state-of-the-art RL algorithms like SAC to solve complex manipulation tasks such as controlling multiarm manipulators~\cite{prianto_2020_PathPlanning}.

There have been many approaches that introduce safety constraints to the exploration process of RL agents.
Simple reward shaping has shown to be insufficient to guarantee safety as it is infeasible for universally deciding between fulfilling the safety constraints and rapidly accomplishing the goal~\cite{achiam_2017_ConstrainedPolicy, ray_2019_BenchmarkingSafe}.
Many recent works use constrained policy optimization techniques \cite{achiam_2017_ConstrainedPolicy, dalal_2018_SafeExploration, ray_2019_BenchmarkingSafe} to reduce the number of safety-critical interactions with the environment.
Although some of these approaches show a very low number of safety constraint violations, none can warrant safety at all times.

In many scenarios, especially when interacting with humans, a very low chance of collision is still unacceptable. 
Therefore, recent works \cite{pham_2018_OptLayerPractical, krasowski_2020_SafeReinforcement, hunt_2021_VerifiablySafe, shao_2021_ReachabilityBasedTrajectory} in the field of provably safe RL (often referred to as shielding, safety layers, sandboxing, or constraining bolts) focus on provable safety guarantees.
The core idea of these methods is to ensure that only safe actions are sent to the environment. 
As a result, unsafe actions are either replaced with safe actions, projected to the safe action space, or entirely prevented by limiting the action space.
Many provably safe RL approaches have been proposed for simple environments such as Atari games~\cite{jansen_2020_SafeReinforcement}, grid worlds~\cite{alshiekh_2018_SafeReinforcement, konighofer_2021_OnlineShielding}, or board games~\cite{degiacomo_2021_FoundationsRestraining}.
Most notably, Hunt\etal~\cite{hunt_2021_VerifiablySafe} exhibited how safe end-to-end learning from image data could be achieved in discrete action spaces.
Unfortunately, these methods rely on a deterministic environment to assure safety, so they are not directly applicable to a complex task like human--robot collaboration or coexistence. 
For the use case of RL for manipulators, Pham\etal~\cite{pham_2018_OptLayerPractical} proposed a differentiable safety layer~(OptLayer) that projects any unsafe action to the closest safe action that satisfies the given constraints. 
Despite promising results, this method comes with two flaws.
First, the OptLayer is only formally correct if the constraints of the underlying quadratic program are precisely fulfilled, and the optimizer finds an exact solution in the given time.
Second, it is difficult to formalize the safety constraints for unknown and complex human motions.
The reachable set-based safety layer of~\cite{krasowski_2020_SafeReinforcement} provides formal safety for autonomous driving highway scenarios by masking out all nonsafe actions. Training an RL agent with this safety layer leads to faster convergence while all collisions can be avoided. 
This method was designed for a discrete action space and is therefore not directly transferable to the manipulator application.
The approach closest to ours is the reachability-based trajectory safeguard of~\cite{shao_2021_ReachabilityBasedTrajectory}. 
They use reachability analysis to predict if the agent could collide with static obstacles on the current trajectory and replan a safe trajectory if necessary.
However, this approach is only designed and tested for static environments.

\subsection{Contributions}
We present a novel safety shield\footnote{Our code and models are publicly available at \url{https://github.com/JakobThumm/safe_rl_manipulators}.} that replaces unsafe actions from the agent with provably safe actions from a high-frequency safe trajectory planner (failsafe planner).
To the best of our knowledge, our method is the first safety measure for RL that provides provable safety for continuous action spaces in high-dimensional state-spaces and hard-to-predict dynamic environments.
Our safety shield can easily be applied to a large variety of manipulation tasks.
Compared to all previously mentioned provably safe RL approaches, we sample from the entire set of actions (instead of a safe subset) and check the safety of the current action during execution at high frequency.
This allows for quick reactions to highly dynamic human motion.
To summarize, our safety shield guarantees that the robot stops before a collision with a human could occur while still allowing the maximal freedom of movement under the given safety constraints.
%The separation of safety checking and decision making allows the RL agent to start long-term actions, although their safety will be fully verified later in an online fashion.
%To make the existing failsafe planner usable in our safety shield, we add a method to repeatedly reach new intermediate goals outputted by the RL agent.
%Furthermore, we adapt the verification of the robot trajectory of previous approaches to better suit the RL setup.

\subsection{Article structure}
\cref{Sec_Preliminaries} summarizes our problem and presents the RL basics and notation used in the article.
Next, \cref{Sec_Methods} presents our proposed safety shield and RL agent.
We then discuss the experimental setup and results in~\cref{Sec_Results} and present our conclusion in~\cref{Sec_Conclusion}.

%% file: preliminaries.tex
\subsection{Problem statement}
\begin{figure}[t]
    \centering
    \vspace{0.17cm} 
    \includegraphics[width=\columnwidth]{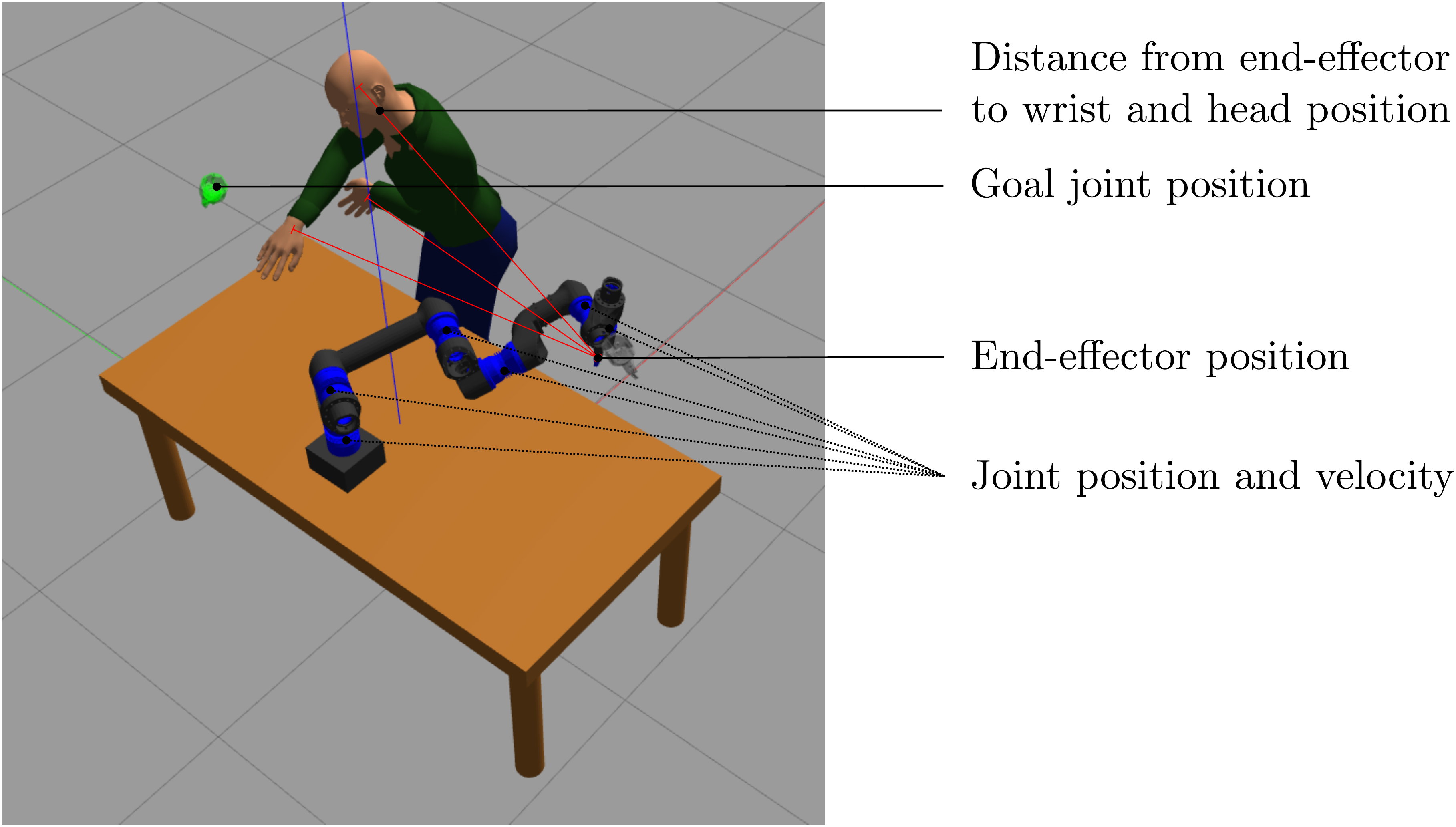}
    \caption{Simulation setup with the robot mounted on a desk and the human performing a task. The provided observations are listed on the right hand side.}
    \label{fig:observations}
\end{figure}

In our RL setting, a six DoF modular robot, as described in~\cite{althoff_2019_EffortlessCreation}, is mounted on a working table, where a human repeatedly performs a task.
In each episode, the agent has to reach a randomized goal joint position, further referred to as \textit{episode goal} $g$, from a fixed initial position and evade the human.
We only consider episode goals that are collision-free with the static environment.
This paper aims to provide formal safety guarantees for nearby humans with unknown motion behavior while consistently reaching the goal.
Hereby, the RL agent receives the following observations: the current joint position and velocity, episode goal, current Cartesian end-effector position, and the relative Cartesian positions of the human wrists and head in relation to the end-effector.
Since the only static obstacle in our scene is the table, we do not add static obstacle information to the observation space.
\cref{fig:observations} shows our setup and illustrates the observations.

\begin{figure*}[ht]
\centering
\vspace{0.17cm} 
\includegraphics[width=0.97\textwidth]{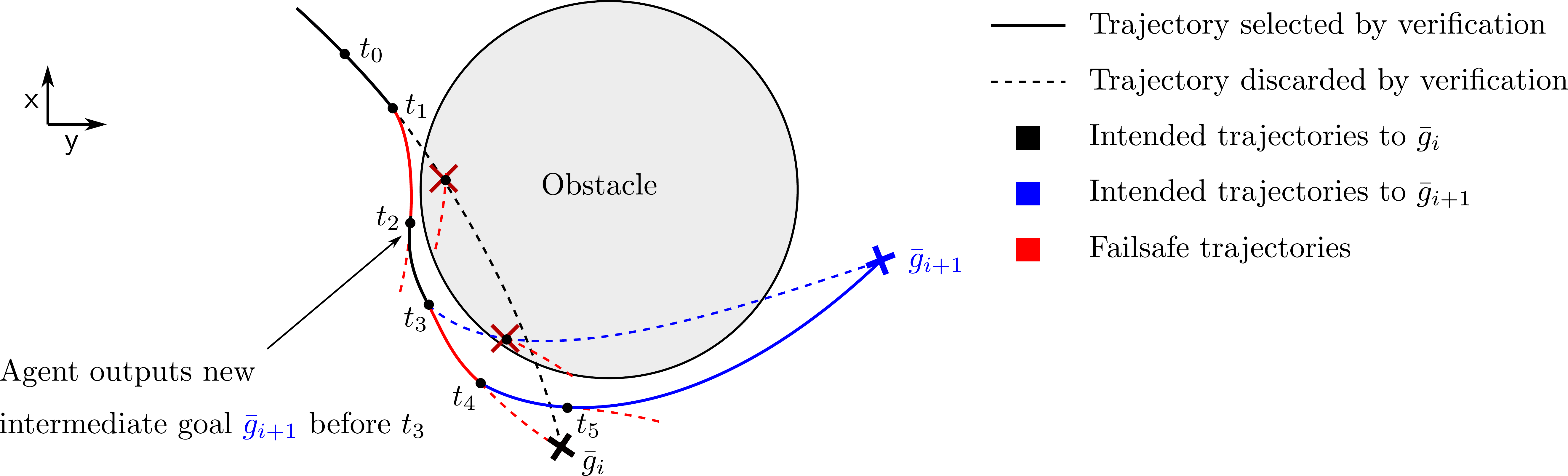}
\caption{Illustration of intended and failsafe trajectories. In time steps $t_0 \dots t_2$, the intermediate goal for the intended trajectories is $\bar{g}_i$.
Before time step $t_0$, a failsafe trajectory is calculated for time step $t_1$. 
In between $t_0$ and $t_1$, an intended trajectory for $t_1$ followed by a failsafe trajectory for $t_2$ is calculated.
The intended trajectory in $t_1$ is not safe, therefore the robot executes the failsafe trajectory.
A new intended trajectory for $t_2$ followed by a failsafe trajectory for $t_3$ is calculated during execution of the failsafe trajectory between $t_1$ and $t_2$.
At $t_2$, the intended trajectory is verified as safe and can be followed.
Before the trajectory calculation for $t_3$, the agent outputs a new intermediate goal $\bar{g}_{i+1}$, so that the new intended trajectory is calculated from $t_3$ to $\bar{g}_{i+1}$. 
Since the intended trajectory to $\bar{g}_{i+1}$ is not verified as safe at $t_3$, the robot follows the previous failsafe trajectory. 
A new intended trajectory from $t_4$ to $\bar{g}_{i+1}$ is calulated followed by a failsafe trajectory.
This intended trajectory is safe, so that the robot switches to it.}
\label{fig:trajectory_calculation}
\end{figure*}
\subsection{RL}
%% Markov decision process
For RL, we use the notation in~\cite{haarnoja_2018_SoftActorCritic} and consider the Markov decision process defined by the tuple $\left( \mathcal{S}, \mathcal{A}, p, r\right)$ with both continuous state space $\mathcal{S}$ and action space $\mathcal{A}$.
For simplicity, we assume the state to be fully observable.
The transition function $p : \mathcal{S} \times \mathcal{S} \times \mathcal{A} \rightarrow \mathbb{R}$ denotes the probability density function of reaching the next state $\boldsymbol{s}_{i+1}$ when choosing action $\boldsymbol{a}_{i}$ in state $\boldsymbol{s}_{i}$.
After each transition, the agent receives a reward from the environment according to the reward function $r : \mathcal{S} \times \mathcal{A} \rightarrow \mathbb{R}$.
The agent learns a stochastic policy $\pi(\boldsymbol{a}_{i}|\boldsymbol{s}_{i})$ for action $\boldsymbol{a}_{i}$ given state $\boldsymbol{s}_{i}$.

%% Soft actor-critic 
The optimal policy is learned using the online off-policy SAC algorithm first presented in~\cite{haarnoja_2018_SoftActorCritic}.
Our specific implementation extends the \texttt{spinningup}~\cite{achiam_2018_SpinningDeep} version of SAC with HER and is further described in~\cref{sec:RL}. 
The goal of SAC is to automatically balance exploration and exploitation by maximizing the accumulated reward and the information content of the policy function.
Therefore, the entropy~$\mathcal{H}$ (see e.g.~\cite[Eq.~4.2]{ziebart_2010_ModelingPurposeful}) of the policy is used as part of the optimization objective (from \cite[Eq.~1]{haarnoja_2018_SoftActorCritic})
\begin{equation}
    J(\pi) =\sum_{t=0}^{T} \mathbb{E}_{\left(\boldsymbol{s}_{i}, \boldsymbol{a}_{i}\right) \sim \rho_{\pi}}\left[r\left(\boldsymbol{s}_{i}, \boldsymbol{a}_{i}\right)+\alpha \mathcal{H}\left(\pi\left(\cdot \mid \boldsymbol{s}_{i}\right)\right)\right].
\end{equation}
%% Hindsight experience replay
In many robotic applications, one encounters the problem of sparse rewards, where the agent needs to get to a random and hard-to-reach goal.
If the agent rarely reaches the goal, almost no transitions with goal encounters are added to the replay buffer.
In order to create more of these goal transitions, \cite{andrychowicz_2017_HindsightExperience} proposed HER with the idea to sample additional transitions with fictional goals, where the goal $g$ of an episode is added to state $\boldsymbol{s}_{i}$, written as $\boldsymbol{s}_{i}||g$. The goal-conditioned transition tuples can be written as $\left(\boldsymbol{s}_{i}||g, \boldsymbol{a}_{i}, r_{i}, \boldsymbol{s}_{i+1}||g\right)$.
For each transition, additional fictional episode goals $\tilde{g}^{h}, \, h = 1 \dots k_{\text{HER}}$ are created from states that were actually reached in the episode of the transition, leading to $k_{\text{HER}}$ (in our case $k_{\text{HER}}=4$) new transitions $\left(\boldsymbol{s}_{i}||\tilde{g}^{h}, \boldsymbol{a}_{i}, \tilde{r}_i^{h}, \boldsymbol{s}_{i+1}||\tilde{g}^{h}\right)$.
By sampling new goals from states that have actually been reached, many successful goal transitions are added.
This method enables us to use a sparse reward, where $r_i=0$ if the goal was reached and $r_i=-1$ otherwise. Since we terminate an episode if a collision occurs, e.g., the human walks into the stationary robot, a negative reward for collisions is not strictly necessary.

%% file: methods.tex
\subsection{Safety shield}
Our proposed safety shield is based on the provably safe trajectory planner, called \textit{failsafe planner} %first presented
in~\cite{beckert_2017_OnlineVerification}. 
The shield functionality is illustrated in~\cref{fig:concept} and further elaborated in this subsection.
Please refer to \cite{beckert_2017_OnlineVerification, pereira_2017_CalculatingHuman, althoff_2019_EffortlessCreation, pereira_2018_OverapproximativeHuman} for a more detailed description of the trajectory planning and verification process.

In our application, the actions of the RL agent are \textit{intermediate goal} joint positions $\bar{g}_i$. 
These intermediate goals $\bar{g}_i$ are not to be confused with the episode goal $g$ or the fictional episode goals for HER $\tilde{g}^h$. 
In each episode, the agent outputs multiple intermediate goals $\bar{g}_i$ to reach the single episode goal $g$. 
The HER goals $\tilde{g}^h$ are only relevant for training the agent.
Furthermore, we would like to emphasize that the RL agent outputs the intermediate goals at a lower frequency than our safety shield is operated.
Each RL action is executed for a user-defined time interval $\Delta T$, or until all joints are within a small $\epsilon$-range of the intermediate goal. 
We refer to this as one \textit{RL step} and use the index $i$.
The user-defined time between two safety shield verification steps is $\Delta t$ with $\Delta t < \Delta T$ and the index $k$ is used.

\paragraph{Long-term and failsafe planning}
This paragraph describes the trajectory planning in our safety shield and is accompanied by a simplified example in~\cref{fig:trajectory_calculation}.
To reach the intermediate goal $\bar{g}_i$, we calculate an \textit{intended trajectory} from the current joint state $\boldsymbol{x}_k$ to $\bar{g}_i$ using a long-term planner.
We guarantee that a failsafe trajectory always exists, which brings the robot to an inherently safe state, i.e., a full stop, because a dynamic obstacle can block the robot's path during the execution of this intended trajectory.
Formal safety can now be guaranteed by induction, where it is assumed that the robot starts in a safe state (stopped).
The first intended trajectory is executed starting from $t_0$ if no collision with a human is possible between $t_0$ and $t_1$ and if a collision-free failsafe trajectory exists starting from $t_1$.
During the execution of a trajectory between $t_k$ and $t_{k+1}$, we calculate a failsafe trajectory starting from $t_{k+2}$.
If this failsafe trajectory cannot be verified as safe, we execute the previously verified failsafe trajectory starting from $t_{k+1}$.
Since each failsafe trajectory ends in a safe state, we can guarantee safety for an infinite time horizon.
Similarly, one time step of the intended trajectory between $t_k$ and $t_{k+1}$ is only executed if it is verified as safe; otherwise, the failsafe trajectory is executed.
In each time step, a new intended and failsafe trajectory is planned.

In our RL scenario, the agent repeatedly updates the intermediate goal.
%This is covered by our safety shield and also exemplary displayed in~\cref{fig:trajectory_calculation}.
If there is a new intermediate goal $\bar{g}_{i+1}$ at time step $t_k$, the intended trajectory is calculated from the state at time $t_{k+1}$ to $\bar{g}_{i+1}$, and a new failsafe trajectory is calculated from the new intended trajectory starting at time $t_{k+2}$.
If the new intended and failsafe trajectories can be verified as safe, the robot changes to the new intended trajectory.
Otherwise, it follows the last verified failsafe trajectory at $t_{k+1}$.

\paragraph{Adaption to manipulator control}
In our use case, the 3D trajectory planning can become computationally expensive, especially in regards to the high-frequency requirement.
Therefore, we do not recompute the intended trajectory in every time step.
Instead, we only compute intended trajectories for every new intermediate goal $\bar{g}_{i+1}$ until an executable one is found and use velocity scaling for the failsafe trajectory as described in~\cite{beckert_2017_OnlineVerification}.
In short, a failsafe trajectory is path-consistent with the intended trajectory and ends in the complete stop of the robot.
We use the synchronous \textit{Type IV} online trajectory planner of~\cite{kroger_2010_OnlineTrajectory} for the intended and the failsafe planner.
Thus, the intended trajectory planning adheres to velocity, acceleration, and jerk limits ($v^{\text{traj}}_{\text{max}}$, $a^{\text{traj}}_{\text{max}}$, and $j^{\text{traj}}_{\text{max}}$), which are gentle for the manipulator joints.
To achieve fast-braking, we increase the acceleration and jerk limits for the failsafe planning ($a^{\text{failsafe}}_{\text{max}}$ and $j^{\text{failsafe}}_{\text{max}}$ ) to the physical limits given by the robot manufacturer.

\paragraph{Verification}
The trajectory verification is based on collision checking of the reachable sets of all possible human motions and the intended robot trajectory.
It is assumed that no human joint moves faster than \SI{2}{m/s} as defined in DIN~EN~ISO~13855:2010~\cite{_2010_SafetyMachinery}, and all human joint positions are reliably measurable within a specified error bound, e.g., by using a motion capture system.
The safety verification also works with other perception methods like light curtains, as presented in~\cite{pereira_2015_SafetyControl}, but the robot's movement would become more conservative.
The entire space a human can occupy within the time interval $\left[t_k, t_{k+b}\right]$ is defined as the reachable occupancy $\Gamma\left(\left[t_{k}, t_{k+b}\right]\right)$, where $b$ is the number of time steps needed to bring the robot to a complete stop, and is calculated using the task space approach presented in~\cite{pereira_2017_CalculatingHuman, althoff_2019_EffortlessCreation}.
%For the verification of a single time step of an intended trajectory $b$ equals \num{1}, and for a failsafe trajectory $b$ is equal to 

% \begin{equation}
%     \Gamma\left(\left[t_{i}, t_{j}\right]\right) \supseteq\left\{\mathcal{F}(\boldsymbol{\xi}) \mid \boldsymbol{\xi} \in \Xi\left(\left[t_{i}, t_{j}\right]\right)\right\},
% \end{equation}
% where $\boldsymbol{\xi}$ is the internal state of the human body, $\mathcal{F}(\boldsymbol{\xi}) \subset \mathbb{R}^{3}$ is the spatial occupancy of the human at a particular state, and $\Xi$ is the set of reachable states starting from an initial state $\boldsymbol{\xi}(t_{0})$.
%Thereby, the joints are assumed to have fixed position, velocity, and acceleration limits, resulting in three occupancy human models $\Gamma_{\text{pos}}$, $\Gamma_{\text{vel}}$, and $\Gamma_{\text{acc}}$.

\begin{figure}[t]
    \centering
    \vspace{0.17cm} 
    \includegraphics[width=0.9\columnwidth]{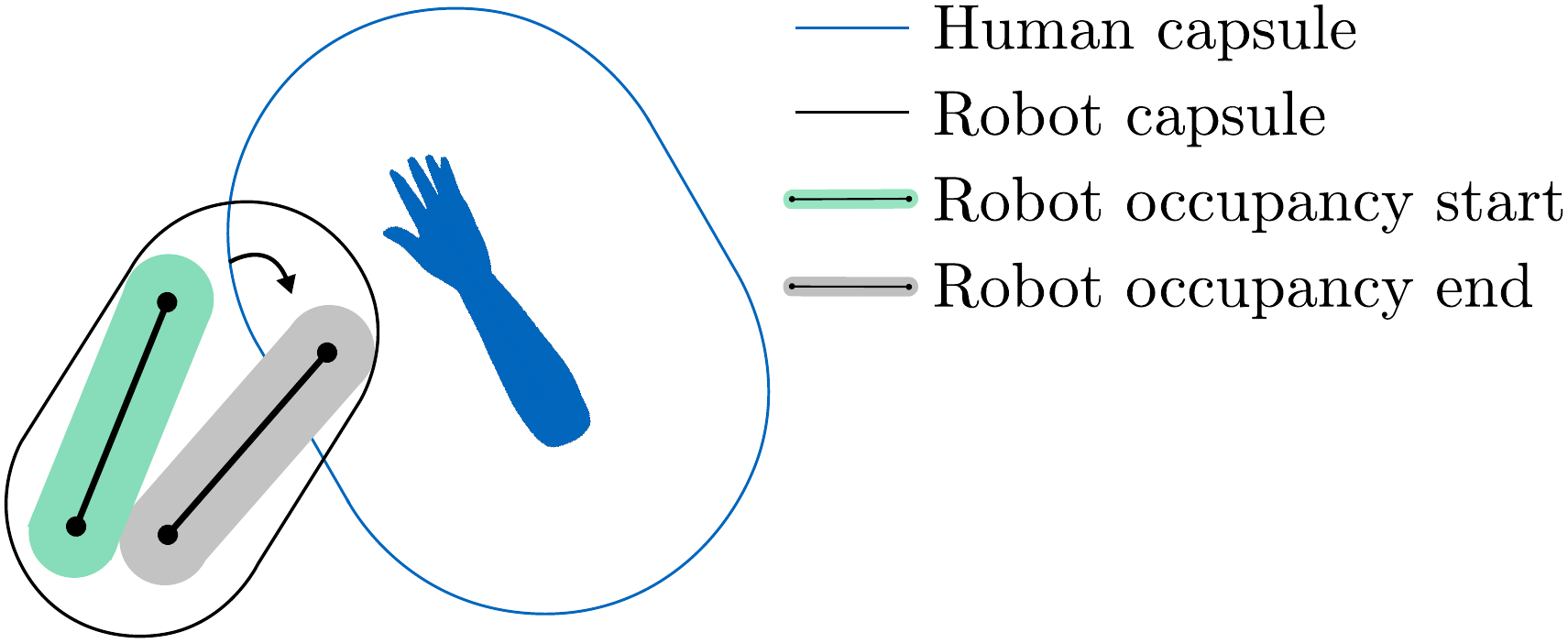}
    \caption{Example of a human--robot capsule pair in collision.}
    \label{fig:safe_capsule}
\end{figure}

The human and robot occupancies are modeled with capsules, as shown in \cref{fig:safe_capsule}, to achieve fast computation times. 
A capsule comprises a cylinder with half-spheres at both ends and is defined by a line segment $l({p}_1, {p}_2)$ with endpoints ${p}_1$ and ${p}_2$ and a radius $r$.
The robot occupancy is described as a set of capsules enclosing each robot link's movement between time $t_k$ and $t_{k+b}$. Hereby, we follow the approach described in~\cite{beckert_2017_OnlineVerification}
%~Section~IV~A 
to guarantee over approximitaion of the link occupancies. 
This approach assumes that the controller follows the desired trajectory exactly.
If additionally deviation occurring due to low-level control should be considered, the approach presented in~\cite{pereira_2015_SafetyControl} can be used.
%All human and robot capsules are over approximative and take modeling and measurement errors into account.
A robot motion is verified as safe if
%, for at least one of the three human occupancy models,
no robot capsule intersects with any human capsule for all times before reaching the resting position.

\subsection{Safe RL}\label{sec:RL}
\begin{figure}[t]
\vspace{0.17cm} 
  \centering
  \begin{minipage}{.98\linewidth}
    \removelatexerror % Nullify \@latex@error
    \begin{algorithm}[H]
      \SetAlgoLined
        \SetKwBlock{Initialize}{Initialize:}{end}
        \SetKw{In}{in}
        \SetKw{KwAnd}{and}
        \SetKw{KwBreak}{break}
        \SetArgSty{textup}
        \Initialize{
        \begin{itemize}
            \item soft actor-critic agent $\mathbf{A}$
            \item environment $\mathbf{E}$
            \item replay buffer $R$ and local buffer $L$
            \item $T_{\text{total}} \gets 0$, $T_{\text{last\_update}} \gets 0$
            \item randomly initialize weights of $\pi_b$
        \end{itemize}
        }
        \For{$j\gets0$ \KwTo $n_{\text{epochs}}$}{
            \For{$l\gets0$ \KwTo $n_{\text{episodes\_per\_epoch}}$}{
                reset $\mathbf{E}$ and clear $L$ \label{algo:training:reset}\;
                \For{$i\gets0$ \KwTo $t_{\text{max\_episode}}$}{
                    \eIf{$T_{\text{total}} \ge k_{\text{start\_steps}}$}{ \label{algo:training:startstep}
                        $a_i \gets \pi_b\left(s_i || g\right)$\;
                    }{
                        $a_i \gets $ random action\;
                    }
                    $\left(s_{i+1}, \, r_i, \, done\right) \gets \mathbf{E}$.step($a_i$) \label{algo:training:step}\;
                    $L\left[i\right] \gets \left(s_i || g, \, a_i, \, r_i, \, s_{i+1} || g\right)$\;
                    $T_{\text{total}} \gets T_{\text{total}} +1$\;
                    \lIf{done}{\KwBreak}
                }
                $l_{episode} \gets i$\;
                \For{$i \gets 0$ \KwTo $l_{episode}-1$}{\label{algo:training:HER}
                    $\left(s_i || g, \, a_i, \, r_i, \, s_{i+1} || g\right) \gets L\left[i\right]$\;
                    $R$.store$\left(s_i || g, \, a_i, \, r_i, \, s_{i+1} || g\right)$\;
                    remaining\_ids $\gets (i+1) \dots (l_{episode}-1)$\;
                    randomly select $k_{\text{HER}}$ random\_ids from remaining\_ids\;
                    \For{id \In random\_ids}{
                        $\left(s_{id} || g, \, \dots \right) \gets L\left[id\right]$\;
                        $\tilde{g} \gets s_{id}$\;
                        $\tilde{r}_i \gets \mathbf{E}$.ComputeReward($s_{i+1} || \tilde{g}$)\;
                        $R$.store$\left(s_i || \tilde{g}, \, a_i, \, \tilde{r}_i, \, s_{i+1} || \tilde{g}\right)$ \label{algo:training:HERend}\;
                    }
                }
                \If{$T_{\text{total}} \ge T_{\text{update\_after}}$ \KwAnd $(T_{\text{total}}-T_{\text{last\_update}}) \ge T_\text{update\_every}$}{\label{algo:training:updateafter}
                    \For{$m\gets0$ \KwTo $T_{\text{total}} - T_{\text{last\_update}}$}{
                        $\mathbf{A}$.Update($R$.sample\_batch())\label{algo:training:updateafterend}\;
                    }
                    $T_{\text{last\_update}} \gets T_{\text{total}}$\;
                }
            }
        }
         \caption{Training procedure}
         \label{algo:training}
    \end{algorithm}
  \end{minipage}
\end{figure}

% In principle, there are two common action space representations for RL for manipulators; either the relative desired joint positions or the desired end-effector position at
% the next RL step. 
% As the robot's movement per RL step is limited, relative positions are preferred over absolute positions.
% Hereby, if $x^i_{t}$ is the position of joint $i$ at RL step $t$, the action $a^i_t \in [-1; 1]$ results in the new target position $x^i_{t+1} = x^i_{t} + a^i_t \Delta x_{\text{max}}$, where $\Delta x_{\text{max}}$ is the maximum position difference per RL step.
% The joint position actions have the advantage of reaching arbitrary end-effector orientations in comparison to the second approach, where only the end-effector position is controlled and the orientation is fixed.
% Furthermore, no inverse kinematics is needed in the joint position control.
% Therefore, we choose the relative joint positions as actions despite the slightly larger action space.
% Though, our safety shield is also easily applicable to end-effector position control with inverse kinematic.

Our policy network has a $\tanh$ activation function for the output layer and thus outputs actions $a_i = [a^0_i, a^1_i, \dots, a^N_i], a^n_i \in [-1; 1]  \forall n \in 1 \dots N$ with $N$ being the number of joints.
We convert these actions to intermediate goal joint positions with $q^n_{\bar{g}, i+1} = q^n_i + a^n_i \Delta q_{\text{max}}$, where $q^n_i$ is the current position of joint $n$ and $\Delta q_{\text{max}}$ is the maximum joint position difference per RL step.
In recent literature, it is more common to use the relative change in 3D position of the robot end-effector as action space and calculate $q^n_{i+1}$ using inverse kinematics (e.g. in \cite{li_2020_PracticalMultiObject}) since it leads to a lower dimensional action space.
However, we use joint positions as actions to have full control over each joint.
Nevertheless, both approaches are equally compatible with our proposed safety shield.
The choice of $\Delta q_{\text{max}}$ depends on the maximum velocity of the robot $v^{\text{traj}}_{\text{max}}$ and the execution time of each RL step $\Delta T$.
Choosing $\Delta q_{\text{max}} \ge v^{\text{traj}}_{\text{max}} \Delta T$ allows the agent to output a new action before the current action is finished. 
This is desirable because each trajectory ends in a stopped state and we do not want the robot to stop after each action.
%Another approach commonly used in literature is to output the Cartesian position of the end-effector as actions and calculate $q^n_{i+1}$ using inverse kinematics.
%Both approaches are equally compatible with our proposed safety shield.
If the robot collides with the static environment in its intermediate goal, a new action is generated randomly with a uniform distribution over the entire action space until a collision-free intermediate goal is determined.
%This resampling of actions is equivalent to a \textit{safety-refined environment} as formally defined in~\cite{hunt_2021_VerifiablySafe}.
An alternative to this method is to project the unsafe action to the safe action space with respect to the static environment. 
However, this is nontrivial in the case of a high-dimensional manipulator.

In our training procedure, which is described in~\cref{algo:training}, an episode is \textit{done} if the episode goal is reached, the robot is in collision, or by timeout when the maximum number of RL steps $T_{\text{max\_episode}}$ is reached.
We consider the episode goal as reached, if $|q^n_{i+1}-q^n_g| < \epsilon_{g}, \forall n \in 1 \dots N$, where $q^n_{i+1}$ is the position of joint $n$ after the environment step, $q^n_g$ is the episode goal position of joint $n$, and $\epsilon_{g}$ is a small user-defined tolerance for reaching the goal.
At the beginning of each episode, the environment is reset (\cref{algo:training}~\cref{algo:training:reset}) by moving the robot back to its initial position and resetting the human animation.
The action of a single RL step is executed in~\cref{algo:training:step} by the function $\mathbf{E}$.step($a_i$).
To achieve a high exploration rate in the beginning of the training, we choose random actions instead of the policy for the first $k_{\text{start\_steps}}$ RL steps (\cref{algo:training:startstep}).
After each episode, HER samples are added to the replay buffer as described in~\cref{algo:training:HER} to~\cref{algo:training:HERend}.
The agent networks are first updated after $T_{\text{update\_after}}$ RL steps to improve exploration at the start of training and then every $T_\text{update\_every}$ RL steps from~\cref{algo:training:updateafter} to~\cref{algo:training:updateafterend}.
All relevant hyperparameters for RL training and operation are given in~\cref{tab:hyperparameters}.

\begin{table}[t]
\centering
\vspace{0.17cm} 
\caption{Hyperparameters}
\begin{tabular}{l c}
\toprule
Parameter & Value \\ \midrule
$\Delta t$ & \SI{4}{ms} \\
$\Delta T$ & \SI{200}{ms} \\
$v^{\text{traj}}_{\text{max}}$ & \SI{2}{rad \per s} \\
$a^{\text{traj}}_{\text{max}}$ & \SI{2}{rad \per s^2} \\
$j^{\text{traj}}_{\text{max}}$ & \SI{15}{rad \per s^3} \\
$a^{\text{failsafe}}_{\text{max}}$ & \SI{10}{rad \per s^2} \\
$j^{\text{failsafe}}_{\text{max}}$ & \SI{400}{rad \per s^3} \\
$\Delta q_{\text{max}}$ & \SI{0.4}{rad}\\[20.5pt]
\bottomrule
\end{tabular}
\,%\quad
\begin{tabular}{l c}
\toprule
Parameter & Value \\ \midrule
Hidden layers (all networks) & \num{3} \\
Hidden units per layer & \num{64} \\
Minibatch size & \num{128} \\
Replay buffer size & $10^6$ \\
Discount factor $\gamma$ & \num{0.99} \\
Entropy tradeoff $\alpha$ & \num{0.2} \\
$k_{\text{start\_steps}}$ & \num{5000} \\
$T_{\text{update\_after}}$ & \num{1000} \\
$T_{\text{update\_every}}$ & \num{200}\\
$n_{\text{epochs}}$ & 200 \\
$n_{\text{episodes\_per\_epoch}}$ & 30
\\ \bottomrule
\end{tabular}
\label{tab:hyperparameters}
\end{table}

\begin{figure*}[t]
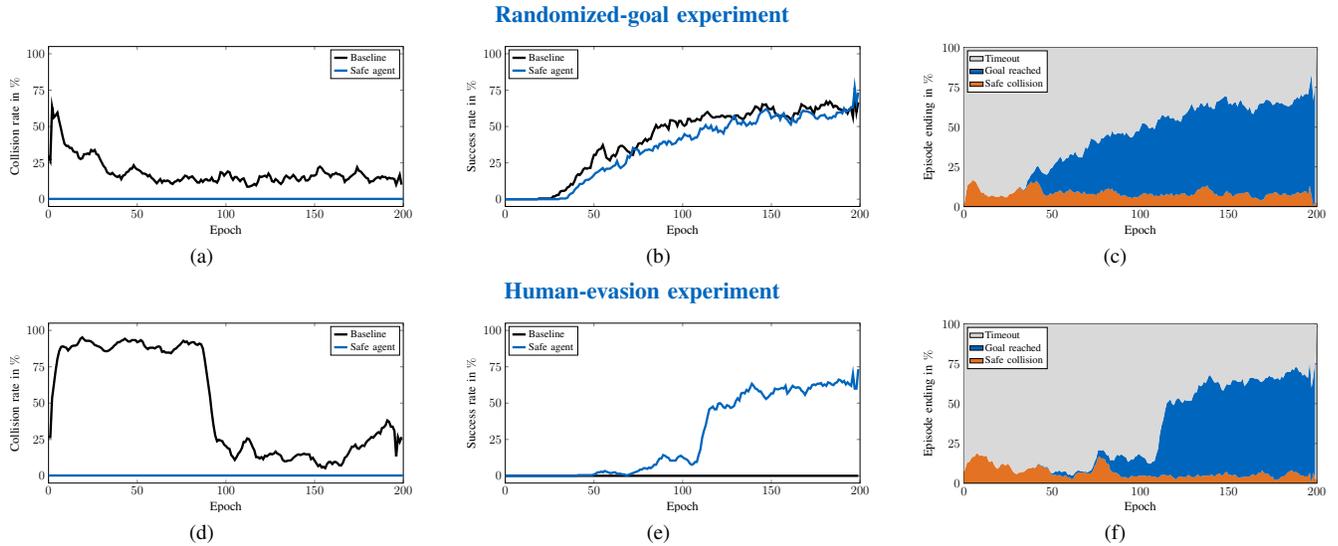

\vspace{0.17cm} 
    \includestandalone[trim=0.0cm 0.0cm 0.0cm 0.0cm, clip, mode=buildnew, width=\linewidth]{tikz/randomized_goal_hor}
    \\[-8pt]
    \subfloat[]{
	\includestandalone[trim=0.2cm 0.2cm 0.2cm 0.0cm, clip, mode=buildnew, width=0.3\linewidth]{tikz/collision_rate_6dof_rand}
	\label{fig:collision_6DoF_rand}
	}
	\subfloat[]{
	%\centering
    \includestandalone[trim=0.2cm 0.2cm 0.2cm 0.0cm, clip, mode=buildnew, width=0.3\linewidth]{tikz/done_rate_6dof_rand}
	\label{fig:goal_reached_6DoF_rand}
	}
	\subfloat[]{
	\includestandalone[trim=0.2cm 0.2cm 0.2cm 0.0cm, clip, mode=buildnew, width=0.3\linewidth]{tikz/episode_ending_safe}
	\label{fig:episode_ending_rand}
	}
	\\[2pt]
    \includestandalone[trim=0.0cm 0.0cm 0.0cm 0.0cm, clip, mode=buildnew, width=\linewidth]{tikz/evade_humans_hor}
    \\[-8pt]
	\subfloat[]{
	\includestandalone[trim=0.2cm 0.2cm 0.2cm 0.0cm, clip, mode=buildnew, width=0.3\linewidth]{tikz/collision_rate_6dof_fixed}
	\label{fig:collision_6DoF_fixed}
	}
	\subfloat[]{
    \includestandalone[trim=0.2cm 0.2cm 0.2cm 0.0cm, clip, mode=buildnew, width=0.3\linewidth]{tikz/done_rate_6dof_fixed}
	\label{fig:goal_reached_6DoF_fixed}
	}
	\subfloat[]{
	\includestandalone[trim=0.2cm 0.2cm 0.2cm 0.0cm, clip, mode=buildnew, width=0.3\linewidth]{tikz/episode_ending_safe_fixed}
	\label{fig:episode_ending_fix}
	}
	\caption{Evaluation of the randomized-goal (upper row) and human-evasion (lower row) experiment. Figures (a) and (d) compare the rate of safety-critical collisions per episode of the baseline and the safe agent. The success rate of both experiments is displayed in (b) and (e). The reason for episode endings using the safe agent are displayed in (c) and (f). Hereby, a safe collision occurs when the human touches the stopped robot.}
\end{figure*}
\subsection{Software structure}
Our software framework is based on \texttt{Gazebo} and \texttt{ROS~Noetic}.
The robot simulation uses the Open Dynamics Engine (ODE) solver with a step size of \SI{1}{ms}.
We choose the ODE QuickStep method with \num{10} solver iterations.
The safety shield is implemented in \CC\footnote{Part of our code is based on the open source reachability analysis tool SaRA (available at https://github.com/Sven-Schepp/SaRA) \cite{schepp_2022_SaRATool}.} to achieve fast calculation times with an average of \SI{0.5}{ms} per verification step\footnote{Run on a ThinkPad P15 Gen 1 with an Intel(R) Core(TM) i7-10750H CPU and \SI{32}{GB} DDR4 RAM.}.
With the safety shield operating at \SI{250}{Hz}, our simulation runs roughly five times faster than real-time.

%For the communication between the RL agent and the \texttt{Gazebo} environment together with the episode handling, we use the \texttt{openai\_ros} package.
%All communication between the Python RL agent and the safety shield is sent over the \texttt{ROS} middleware.

%% file: tikz/randomized_goal_hor.tex
\begin{tikzpicture}
\node [text=TUMblue, text width=20cm, align=center] {\textbf{Randomized-goal experiment}}; 
\end{tikzpicture}

%% file: tikz/collision_rate_6dof_rand.tex
\begin{tikzpicture}
\tikzstyle{every node}=[style={scale=1, transform shape}]
\begin{axis}[
width=\textwidth,
height=0.5\textwidth,
every tick label/.append style={font=\large},
label style={font=\large},
%axis lines=left,
%scaled ticks=false,
ymin=-5, ymax=105,
xmin=0, xmax=200,
xtick distance = 50,
ytick distance = 25,
x tick label style={
	/pgf/number format/.cd,
	fixed,
	fixed zerofill,
	precision=0,
	/tikz/.cd
},
ylabel={Collision rate in $\%$},
xlabel={Epoch},
legend style={at={(0.99, 0.98)},anchor=north east},
legend cell align={left},
nodes={scale=1.35, transform shape},
%legend image post style={mark=*}
]

\pgfplotstableread[col sep=tab]{tikz/data/safe_sac_her_all_rand_6DoF_new_smoothed.txt}\resultSafe;
\pgfplotstableread[col sep=tab]{tikz/data/unsafe_sac_her_all_rand_6DoF_new_smoothed.txt}\resultUnsafe;

% \addplot [
% mark=None,
% line width=1pt,
% color=TUMblue,
% solid
% ] 
% table[
% x = Epoch, 
% y expr=\thisrow{Collided}*100
% ] 
% {\resultSafe};
% \addlegendentry{safe agent}

\addplot [
mark=None,
line width=3pt,
color=black,
solid,
] 
table[
x = Epoch, 
y expr=\thisrow{CriticallyCollided}*100
] 
{\resultUnsafe};
\addlegendentry{Baseline}

\addplot [
mark=None,
line width=1pt,
color=TUMblue,
solid,
line width=3pt,
domain = 0:200
] {0*x+0.2};
\addlegendentry{Safe agent}

% \addplot [
% mark=None,
% line width=1pt,
% color=black,
% solid
% ] 
% table[
% x = Epoch, 
% y expr=\thisrow{Collided}*100
% ] 
% {\resultUnsafe};
% \addlegendentry{Baseline}

\end{axis}
\end{tikzpicture}

%% file: tikz/done_rate_6dof_rand.tex
\begin{tikzpicture}
\tikzstyle{every node}=[style={scale=1, transform shape}]
%data/experimental_uncertainty_prediction/temporal_prediction_hist_CV.csv
\begin{axis}[
width=\textwidth,
height=0.5\textwidth,
every tick label/.append style={font=\large},
label style={font=\large},
%axis lines=left,
%scaled ticks=false,
ymin=-5, ymax=105,
xmin=0, xmax=200,
xtick distance = 50,
ytick distance = 25,
x tick label style={
	/pgf/number format/.cd,
	fixed,
	fixed zerofill,
	precision=0,
	/tikz/.cd
},
ylabel={Success rate in $\%$},
xlabel={Epoch},
legend style={at={(0.01, 0.98)},anchor=north west},
legend cell align={left},
nodes={scale=1.35, transform shape},
]

\pgfplotstableread[col sep=tab]{tikz/data/safe_sac_her_all_rand_6DoF_new_smoothed.txt}\resultSafe;
\pgfplotstableread[col sep=tab]{tikz/data/unsafe_sac_her_all_rand_6DoF_new_smoothed.txt}\resultUnsafe;

\addplot [
mark=None,
line width=3pt,
color=black,
solid
] 
table[
x = Epoch, 
y expr=\thisrow{GoalReached}*100
] 
{\resultUnsafe};
\addlegendentry{Baseline}

\addplot [
mark=None,
line width=3pt,
color=TUMblue,
solid
] 
table[
x = Epoch, 
y expr=\thisrow{GoalReached}*100
] 
{\resultSafe};
\addlegendentry{Safe agent}

\end{axis}
\end{tikzpicture}

%% file: tikz/episode_ending_safe.tex
\begin{tikzpicture}
\tikzstyle{every node}=[style={scale=1, transform shape}]
\begin{axis}[
width=\textwidth,
height=0.5\textwidth,
every tick label/.append style={font=\large},
label style={font=\large},
%axis lines=left,
%scaled ticks=false,
ymin=0, ymax=100,
xmin=0, xmax=200,
xtick distance = 50,
ytick distance = 25,
x tick label style={
	/pgf/number format/.cd,
	fixed,
	fixed zerofill,
	precision=0,
	/tikz/.cd
},
ylabel={Episode ending in $\%$},
xlabel={Epoch},
legend style={at={(0.01, 0.98)},anchor=north west},
legend cell align={left},
nodes={scale=1.35, transform shape},
]

\pgfplotstableread[col sep=tab]{tikz/data/safe_sac_her_all_rand_6DoF_new_smoothed.txt}\resultSafe;

% \addplot [
% mark=None,
% line width=1pt,
% color=TUMblue,
% solid
% ] 
% table[
% x = Epoch, 
% y expr=\thisrow{Collided}*100
% ] 
% {\resultSafe};
% \addlegendentry{safe agent}

\addlegendimage{area legend, fill = mygray}
\addlegendentry{Timeout}

\addlegendimage{area legend, fill = TUMblue}
\addlegendentry{Goal reached}

\addlegendimage{area legend, fill = TUMorange}
\addlegendentry{Safe collision}

\addplot [
mark=None,
line width=0pt,
name path=A,
domain = 0:200
] {0*x+0};

\addplot [
mark=None,
line width=0pt,
opacity=0,
name path=B
] 
table[
x = Epoch, 
y expr=\thisrow{Collided}*100
] 
{\resultSafe};

\addplot [
mark=None,
line width=0pt,
opacity=0,
name path=C
] 
table[
x = Epoch, 
y expr=(\thisrow{GoalReached}) + \thisrow{Collided})*100
] 
{\resultSafe};

\addplot [
mark=None,
line width=0pt,
opacity=0,
name path=D,
domain = 0:200
] {0*x+100};

\addplot [
color=TUMorange
] 
fill between [
of=A and B];

\addplot [
color=TUMblue
] 
fill between [
of=B and C];

\addplot [
color=mygray
] 
fill between [
of=C and D];

% \addplot [
% mark=None,
% line width=1pt,
% color=black,
% solid
% ] 
% table[
% x = Epoch, 
% y expr=\thisrow{Collided}*100
% ] 
% {\resultUnsafe};
% \addlegendentry{Baseline}

\end{axis}
\end{tikzpicture}

%% file: tikz/evade_humans_hor.tex
\begin{tikzpicture}
\node [text=TUMblue, text width=20cm, align=center] {\textbf{Human-evasion experiment}}; 
\end{tikzpicture}

%% file: tikz/collision_rate_6dof_fixed.tex
\begin{tikzpicture}
\tikzstyle{every node}=[style={scale=1, transform shape}]
\begin{axis}[
width=\textwidth,
height=0.5\textwidth,
every tick label/.append style={font=\large},
label style={font=\large},
%axis lines=left,
%scaled ticks=false,
ymin=-5, ymax=105,
xmin=0, xmax=200,
xtick distance = 50,
ytick distance = 25,
x tick label style={
	/pgf/number format/.cd,
	fixed,
	fixed zerofill,
	precision=0,
	/tikz/.cd
},
ylabel={Collision rate in $\%$},
xlabel={Epoch},
legend style={at={(0.99, 0.98)},anchor=north east},
legend cell align={left},
nodes={scale=1.35, transform shape},
]

\pgfplotstableread[col sep=tab]{tikz/data/safe_sac_her_fixed_6DoF_smoothed.txt}\resultSafe;
\pgfplotstableread[col sep=tab]{tikz/data/unsafe_sac_her_fixed_6DoF_smoothed.txt}\resultUnsafe;

\addplot [
mark=None,
line width=3pt,
color=black,
solid,
] 
table[
x = Epoch, 
y expr=\thisrow{CriticallyCollided}*100
] 
{\resultUnsafe};
\addlegendentry{Baseline}

\addplot [
mark=None,
line width=1pt,
color=TUMblue,
solid,
line width=3pt,
domain = 0:200
] {0*x+0.2};
\addlegendentry{Safe agent}

\end{axis}
\end{tikzpicture}

%% file: tikz/done_rate_6dof_fixed.tex
\begin{tikzpicture}
\tikzstyle{every node}=[style={scale=1, transform shape}]
%data/experimental_uncertainty_prediction/temporal_prediction_hist_CV.csv
\begin{axis}[
width=\textwidth,
height=0.5\textwidth,
every tick label/.append style={font=\large},
label style={font=\large},
%axis lines=left,
%scaled ticks=false,
ymin=-5, ymax=105,
xmin=0, xmax=200,
xtick distance = 50,
ytick distance = 25,
x tick label style={
	/pgf/number format/.cd,
	fixed,
	fixed zerofill,
	precision=0,
	/tikz/.cd
},
ylabel={Success rate in $\%$},
xlabel={Epoch},
legend style={at={(0.01, 0.98)},anchor=north west},
legend cell align={left},
nodes={scale=1.35, transform shape},
]

\pgfplotstableread[col sep=tab]{tikz/data/safe_sac_her_fixed_6DoF_smoothed.txt}\resultSafe;
\pgfplotstableread[col sep=tab]{tikz/data/unsafe_sac_her_fixed_6DoF_smoothed.txt}\resultUnsafe;

\addplot [
mark=None,
line width=3pt,
color=black,
solid
] 
table[
x = Epoch, 
y expr=\thisrow{GoalReached}*100
] 
{\resultUnsafe};
\addlegendentry{Baseline}

\addplot [
mark=None,
line width=3pt,
color=TUMblue,
solid
] 
table[
x = Epoch, 
y expr=\thisrow{GoalReached}*100
] 
{\resultSafe};
\addlegendentry{Safe agent}

\end{axis}
\end{tikzpicture}

%% file: tikz/episode_ending_safe_fixed.tex
\begin{tikzpicture}
\tikzstyle{every node}=[style={scale=1, transform shape}]
\begin{axis}[
width=\textwidth,
height=0.5\textwidth,
every tick label/.append style={font=\large},
label style={font=\large},
%axis lines=left,
%scaled ticks=false,
ymin=0, ymax=100,
xmin=0, xmax=200,
xtick distance = 50,
ytick distance = 25,
x tick label style={
	/pgf/number format/.cd,
	fixed,
	fixed zerofill,
	precision=0,
	/tikz/.cd
},
ylabel={Episode ending in $\%$},
xlabel={Epoch},
legend style={at={(0.01, 0.98)},anchor=north west},
legend cell align={left},
nodes={scale=1.35, transform shape},
]

\pgfplotstableread[col sep=tab]{tikz/data/safe_sac_her_fixed_6DoF_smoothed.txt}\resultSafe;

% \addplot [
% mark=None,
% line width=1pt,
% color=TUMblue,
% solid
% ] 
% table[
% x = Epoch, 
% y expr=\thisrow{Collided}*100
% ] 
% {\resultSafe};
% \addlegendentry{safe agent}

\addlegendimage{area legend, fill = mygray}
\addlegendentry{Timeout}

\addlegendimage{area legend, fill = TUMblue}
\addlegendentry{Goal reached}

\addlegendimage{area legend, fill = TUMorange}
\addlegendentry{Safe collision}

\addplot [
mark=None,
line width=0pt,
name path=A,
domain = 0:200
] {0*x+0};

\addplot [
mark=None,
line width=0pt,
opacity=0,
name path=B
] 
table[
x = Epoch, 
y expr=\thisrow{Collided}*100
] 
{\resultSafe};

\addplot [
mark=None,
line width=0pt,
opacity=0,
name path=C
] 
table[
x = Epoch, 
y expr=(\thisrow{GoalReached}) + \thisrow{Collided})*100
] 
{\resultSafe};

\addplot [
mark=None,
line width=0pt,
opacity=0,
name path=D,
domain = 0:200
] {0*x+100};

\addplot [
color=TUMorange
] 
fill between [
of=A and B];

\addplot [
color=TUMblue
] 
fill between [
of=B and C];

\addplot [
color=mygray
] 
fill between [
of=C and D];

% \addplot [
% mark=None,
% line width=1pt,
% color=black,
% solid
% ] 
% table[
% x = Epoch, 
% y expr=\thisrow{Collided}*100
% ] 
% {\resultUnsafe};
% \addlegendentry{Baseline}

\end{axis}
\end{tikzpicture}

%% file: results.tex
% \begin{figure*}[ht!]
%     \subfloat{
% 	\includestandalone[trim=0.2cm 0.2cm 0.2cm 0.0cm, clip, mode=buildnew, width=\textwidth]{tikz/collision_rate_6dof_fixed}
% 	\caption{}%The percentage of safety-critical collisions per episode in the randomized goal experiment. The baseline causes collisions in approximately \SI{20}{\%} of all episodes, whereas the agent with safety shield is fully safe.}
% 	\label{fig:collision_6DoF_fixed}
% 	}
% 	\subfloat{
% 	%\centering
%     \includestandalone[trim=0.2cm 0.2cm 0.2cm 0.0cm, clip, mode=buildnew, width=\textwidth]{tikz/done_rate_6dof_fixed}
% 	\caption{}%The percentage of reached goals per episode in the randomized goal experiment. Both agents learn in a similar rate.}
% 	\label{fig:goal_reached_6DoF_fixed}
% 	}
% 	\subfloat{
% 	\includestandalone[trim=0.2cm 0.2cm 0.2cm 0.0cm, clip, mode=buildnew, width=\textwidth]{tikz/episode_ending_safe_fixed}
% 	\caption{}%The reason for episode endings using the safe agent  in the randomized goal experiment. A safe collision occurs when the human touches the stopped robot.}
% 	\label{fig:episode_ending_fix}
% 	}
% 	\caption{Evaluation of the human-evasion experiment. (a) compares the rate of safety-critical collisions per episode of the baseline and the safe agent. The success rate is displayed in (b). The reason for episode endings using the safe agent are displayed in (c). Hereby, a safe collision occurs when the human touches the stopped robot.}
% \end{figure*}

\subsection{Experimental setup}
We conduct two different experiments. 
In the first \textit{randomized-goal experiment}, the episode goal is randomly and uniformly sampled over the entire joint space, and the start state is fixed with all joints at their zero position.
Notably, the human may block the episode goal. 
For the second \textit{human-evasion experiment}, we manually define the start and episode goal position so that the robot collides with the human if it takes the shortest path.
The agent must learn to evade the human by taking a longer path or waiting until the human moves away from the table.
In this scenario, the episode goal is only slightly randomized so that the human never blocks the goal state.
The two experiments are also shown in our accompanying video\footnote{Available at https://youtu.be/Lzrs2HQUIOc}.
%Additionally, we set up a custom experiment for the 6 DoF, where the start and goal are placed in such a way, that the fastest path to the goal goes through the human workspace.
%Here, the robot has to learn to avoid the human to reach the goal.
%In this setting, the goal end-effector position is bound to lie in a box of size \SI{0.4 x 0.4 x 0.4}{\meter}.

For the simulation, the human motion is taken from real-world motion capture data provided by CMU\footnote{Data publicly available at http://mocap.cs.cmu.edu/}.
Currently, we are using animation \texttt{62\_01}, containing a human walking to the working table, where they perform a wrenching action.
This motion pattern is diverse and complex and can effectively validate our concept. 
To prevent overfitting to the motion data, we randomize the $\mathsf{x}$- and $\mathsf{y}$-position and start time of the human animation in each episode uniformly in the range of \SIrange{-0.2}{0.2}{\meter} and \SIrange{0}{1}{\second}.

In all experiments, the maximum episode length is set to \num{100} RL steps per episode to give the robot adequate time to move around the human to any goal.
Since an episode may end earlier due to collisions and successful goal achievements, the number of RL steps per episode varies.
Each training consists of \num{200} epochs with \num{30} episodes per epoch, leading to a maximum of \num{600000}~RL steps.

\subsection{Randomized-goal experiment}
The following discussion refers to the agent trained excluding and including a safety shield as \textit{baseline} and \textit{safe agent}, respectively. 
We trained the baseline agent with the same hyperparameters as the safe agent to achieve good comparability. 
In order to improve the performance of the baseline agent, we also tried to operate it at higher frequencies (\SI{50}{Hz} and \SI{250}{Hz}) and provide it with a high negative reward for collisions.
Unfortunately, none of the measures lead to successful training improvements, so we omit these results.
First, we want to highlight the results for completely randomized goal positions.
The collision rate comparison in~\cref{fig:collision_6DoF_rand} with and without a safety shield clearly shows the advantage of our proposed method.
The baseline agent causes a safety-critical collision in approximately~\SI{20}{\%} of all episodes.
Therefore, the baseline is too dangerous to be deployed in a real-world environment, especially at the beginning of the training.
In contrast, the safety shield successfully prevents all safety-critical collisions.
Both agents show a similar success rate in~\cref{fig:goal_reached_6DoF_rand}.
Interestingly, the baseline agent learns slightly faster than the safe agent despite its high collision rate.
At the end of the training, both agents have approximately the same success rate of~\SI{65}{\%}.
\cref{fig:episode_ending_rand} evaluates how the remaining~\SI{35}{\%} of episodes end for the safe agent. 
Approximately~\SI{10}{\%} of episodes end in a safe collision caused by the static human animation walking into the stationary robot.
The remaining~\SI{25}{\%} of episodes ended by exceeding $T_{\text{max\_episode}}$.
A manual investigation of the unsuccessful cases showed that most of the goals were not reachable because humans blocked the path to the goal.

\subsection{Human-evasion experiment}
The human-evasion experiment highlights another advantage of our proposed method.
Since the human blocks the optimal path, the baseline agent collides with the human in almost every episode in the first half of the training as shown in~\cref{fig:collision_6DoF_fixed}. 
In the second half of training, the baseline agent learned to avoid collisions with the human more frequently, but still collides in approximately~\SI{25}{\%} of all episodes.
Again, our safety shield prevents all safety-critical collisions.
\cref{fig:goal_reached_6DoF_fixed} shows that the baseline agent cannot reach the goal because it does not learn how to evade the human.
However, the safe agent learns to move around the human to reach the goal effectively.
%We believe that the baseline agent could give better results if methods like behavior cloning would be employed to show the agent an initial path to the goal.

%% file: conclusion.tex
Our results clearly show that our proposed safety shield is an effective method to avoid collisions with humans in the working environment.
Contrary to existing RL methods for safe manipulator control, our safety shield provides formal safety guarantees in highly dynamic and prior unknown human environments.
The human-evasion experiment also demonstrated that our safety shield could significantly impact the training success in scenarios with a high likelihood of a collision.
%The reduced number of collisions naturally lead to a higher number of goals reached in complex scenarios.
We believe that our method enables the high-level control of manipulators with RL agents in real-world human environments while providing the necessary safety guarantees to any human operator.
Our subsequent goals are to refine the RL agent further and train it on a more complex set of real-world human motions.
Finally, we plan to test our safe RL agent on actual robots and perform tasks in a human working environment.